# Skin Segmentation based Elastic Bunch Graph Matching for efficient multiple Face Recognition

Sayantan Sarkar

Department of Electrical Engineering, NIT Rourkela,
sayantansarkar24@gmail.com

**Abstract.** This paper is aimed at developing and combining different algorithms for face detection and face recognition to generate an efficient mechanism that can detect and recognize the facial regions of input image.

For the detection of face from complex region, skin segmentation isolates the face-like regions in a complex image and following operations of morphology and template matching rejects false matches to extract facial region.

For the recognition of the face, the image database is now converted into a database of facial segments. Hence, implementing the technique of Elastic Bunch Graph matching (EBGM) after skin segmentation generates Face Bunch Graphs that acutely represents the features of an individual face enhances the quality of the training set. This increases the matching probability significantly.

**Keywords:** Elastic Bunch Graph Matching, Skin Segmentation, Gabor Wavelets, Graph Similarity, Template Matching, Face Bunch Graph, Face Recognition Database of University of Essex

## 1  Introduction

Humans inherently use faces to recognize individuals and now, advancements in computing capabilities over the past few decades enable similar recognitions automatically [5]. Face recognition algorithms have, over the years, developed from simple geometric models to complex mathematical representations and sophisticated vector matching processes. [15]

Today, face recognition is actively being used to minimize passport fraud, support law enforcement agencies, identify missing children and combat identity theft.

Typically, the algorithms for face detection and recognition fall under any one of the following broad categories [19] –

1. **Knowledge-based methods:** encode what makes a typical face, e.g., the association between facial features.
2. **Feature-invariant approaches:** aim to find structure features of a face that do not change even when pose, viewpoint or lighting conditions vary(PCA).[17]
3. **Template matching:** comparison with several stored standard patterns to describe the face as a whole or the facial features separately.
4. **Appearance-based methods:** the models are learned from a set of training images that capture the representative variability of faces.

## 2    Approach

As all the algorithms for facial recognition is based on certain set of specified features or template matching, a raw image input leads to over identification of features from the background region.

This over identified features leads to garbage values that alter the matching criteria leading to under identification or false identification.

To minimize such errors, a two-step approach is adopted by us that initially in the first step detects the facial region as the foreground and rejects the rest of the image segment as background. In the second step the facial recognition algorithm is executed only for the foreground region. Hence, the over identification is curtailed, increasing the efficiency of the algorithm with higher ratio of correct identification.

## 3    Image Enhancement

### 3.1    Contrast Stretching

The variance of the image is calculated and checked against a cut threshold, such that the non-uniform light regions get corrected. The contrast enhancement of the image may be done globally or adaptively. The contrast stretching function used is a simple piecewise linear function, although sigmoid functions may also be used.

### 3.2    Color Space Conversion

The standard input images are generally in RGB color space, which is also known as the visual color space. But in this Cartesian representation of the color space, the chroma and the luma components for each pixel is non-distinguishable. Hence, the color space is for each pixel in the image is converted to another popular color space (HSV color space).

## 4    Skin Segmentation

As the initial RGB image is already converted into HSV space after proper contrast stretching to remove luminous variations and dependence we can only focus on the hue and the saturation component [10]. These 'H' and 'S' color component are plotted, and averaged to give two final histograms. The histograms shows that the 'H' and 'S' components for faces are properly clustered.

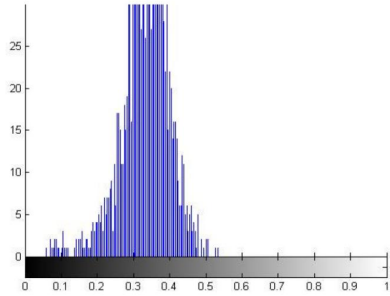

**Fig. 1.** Example Histogram of 'S' component of Training Facial image

In our analysis the highest point of the histograms were considered the Mean Value with a Variance of 50%. The suitable Gaussian Curve corresponding to the extracted data is then superimposed on the original histograms to threshold the image into two segments. The intra Gaussian segment corresponds to the skin segment.

### 4.1 Morphological Noise Removal

After the rejection of the non-skin region, the image still remains noisy due to stray pixels having skin chroma or non-facial skin regions that need to be cleared up [6]. The sequence of steps is further described as follows:

1. As the noise removal is intensity-based, the skin segmented image is converted to gray scale and threshold at a low threshold for background.
2. Morphological opening is performed using a small window convolution to remove the tiny skin segments, not classified as facial region and repeated with 17X17 window after filling intra facial holes.

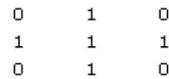

**Fig. 2.** Structuring Window

### 4.2 Facial Region Classification

An image can contain skin region apart from the facial skin region. The skin segmented image is threshold on the basis of skin chroma, which makes the facial components such as eyes and nose to be threshold in the background as they do not match the skin chroma. They show up as 'holes' after proper binarizing via thresholding. The Euler numbers [14] for these binarized regions are then calculated.

An adaptive system is used to produce the threshold for each connected area. If there is a large spread and the ratio of mean to standard deviation is high, the threshold is set to a fraction of the mean. This stops darker faces from splitting up into many connected regions after thresholding.

After computation of the Euler number e. If e < 0 (i.e. less than two holes) the region is rejected. This is based on the fact that the face has at least two holes due to the presence of the eyes.

### 4.3 Template Matching

The Euler Number criteria are not limiting to facial segments. To remove this limitation average template is constructed manually. This template was convolved repeatedly with the image and the maximum peaks were extracted until the peak failed to exceed a calculated threshold value. The coordinates of the center of the bounding boxes which matched the most with this average template were stored in an array.

## 5 Detection

The primary motive of a good facial recognition algorithm is to correlate between a facial segment and a pre-identified facial database. But the performance criteria predominately depend on three factors: 1> Robustness 2> Speed 3> Hit Rate.

In our method the robustness and the speed of the algorithm is improved using the first step of the two step process, which thresholds the irrelevant back ground data. The input to the recognition algorithm is the facial segments extracted from the input images, which reduces the chances of false detection due to excess data points.

### 5.1 Elastic Bunch Graph Matching (EBGM)

The main idea of the EBGM is to design an algorithm that can comparatively analyze two images and find a quantitative value for the similarity between them [7].

The facial image in each bounding box is assumed to be threshold by skin segmentation such that only the skin region of the facial region is identified.

In our current analysis of the algorithm, we are assuming that the matching system is to be used as an integration of user identification systems, where due to the detection algorithm the facial segment is extracted and normalized with reference to a standard size and standard contrast. Hence, we assume that the distortions due to Position and Size are not present in out facial segment database.

So the only variation/ distortion are:

1. **Expression Distortions**: The facial expression of the person in the facial image is prone to change in every image hence cannot match the base database image.
2. **Pose Distortions**: The position of the person with respect to the camera can vary a little in the 2-D plane of the image but not over a large range.

The basic object representation used in case of EBGM is a graph. This graph is used to represent a particular face segment, is therefore a 'facial graph'.

The facial graph is labeled with nodes and edges. In turn the nodes are identified with wavelet responses of local jets and the edges are identified by the length of the edge [3].

1. **Node:** It is a point in the 2-D facial graph of the image that signifies the phase and the magnitude of the wavelet response of the facial image in the locality of the node point.
2. **Edge**: It is a line that connects the nodes. Every two node in the graph is interconnected with an edge, which is represented with the magnitude of the length of the edge.

### 5.2    Create Bunch Graphs

**Gabor Wavelets**

For detection referring the earlier work of Wiscott[18]. Gabor wavelets generated using Gabor Wavelet Transform is applied as Gabor filters to wavelet space [11]. They are predominately edge detection filter that assigns magnitude and phase depending on edge directions and intensity in varying dilation and rotational values. They are mainly implemented by convolving a function with the image to generate Gabor Space [2].For a set of degree of rotations and dilations a set of Gabor kernels are generated. These kernels will extract the 'jets' from the image. [18]

$$J_j(\vec{x}) = \int I(\vec{x}')\psi_j(\vec{x} - \vec{x}')d^2\vec{x}' \quad (1)$$

Convolution with Gabor Kernels to generate wavelet transformed image

$$\psi_j(\vec{x}) = \frac{k_j^2}{\sigma^2}\exp\left(-\frac{k_j^2 x^2}{2\sigma^2}\right)\left[exp(i\vec{k_j}\vec{x}) - \exp\left(-\frac{\sigma^2}{2}\right)\right] \quad (2)$$

Family of Gabor Kernels for *j* varying from 0 to 39

Here $k_j$ is the wave vector that is restricted by the Gaussian envelope function. For our calculations, 5 different set of frequencies for index $v$ = 0, 1…4 and 8 sets of orientation directions µ= 0, 1, 2….7 are taken [18].

$$\vec{k_j} = \begin{pmatrix} k_{jx} \\ k_{jy} \end{pmatrix} = \begin{pmatrix} k_v \cos\varphi_\mu \\ k_v \sin\varphi_\mu \end{pmatrix}, k_v = 2^{-\frac{v+2}{2}}\pi, \varphi_\mu = \mu\frac{\pi}{8}, \quad (3)$$

Wave vector *kj* for *j* varying from 1 to 39

The width $\frac{\sigma}{k}$ is Gaussian controlled with $\sigma = 2\pi$. The preference of Gabor wavelet transform over normal edge detection and analysis is evident in this case as Gabor filters are much more robust to the data format of biological relevance which in this case is facial segments. Also the robustness is defined as the result of the transform is not susceptible to variation brightness when the Gabor wavelets are considered DC-free [18].

If these jets are normalized then the Gabor wavelet transform is set immune to contrast variations. Still rotation and translation to a small degree does not affect the magnitude of jets but result in high phase variations.

The phase can be used to calculate the degree of displacement between two images and compensate this distortion. [13][12]

**Jet Similarity**

These phase values have its set of importance in feature matching [9] as:

1. Similar magnitude patterns can be discriminated
2. The phase variation is a measure for accurate jet localization of a feature point.

In order to stabilize the phase sensitive similarity function, the compensation factor needs to be subtracted that nullifies the phase variation in nearby pixel points. For the compensation factor it is assumed that the jets compared in the similarity function belongs to nearby point hence have a small displacement between them [18]. Thus a small relative displacement $d$ is implemented to generate following phase sensitive similarity function,

$$S_\phi(J,J') = \frac{\sum_j a_j a'_j \cos(\phi_j - \phi'_j - \vec{d}\vec{k_j})}{\sqrt{\sum_j a_j^2 \sum_j a'^2_j}} \quad (4)$$

Similarity Function $S$ for jets including phase

For displacement factor $d$, the Taylor series expansion is further solved to get reduced to:

$$\vec{d}(J,J') = \begin{pmatrix} d_x \\ d_y \end{pmatrix} = \frac{1}{\Gamma_{xx}\Gamma_{yy} - \Gamma_{xy}\Gamma_{yx}} X \begin{pmatrix} \Gamma_{yy} & -\Gamma_{yx} \\ -\Gamma_{xy} & \Gamma_{xx} \end{pmatrix} \begin{pmatrix} \phi_x \\ \phi_y \end{pmatrix} \quad (5)$$

$$\phi_x = \sum_j a_j a'_j k_{jx}(\phi_j - \phi'_j)$$

$$\Gamma_{xy} = \sum_j a_j a'_j k_{jx} k_{jy}$$

Displacement vector

This equation gives the displacement factor between the jets considering they belong to two neighboring pixel points in the locality [18]. Thus the range of the displacement that can be calculated with it extends to half the wavelength of highest frequency kernel that can be up to 8 pixels for high frequencies. [4][16]

**Face Graph Representation**

For face graph generation from facial images, a set of 'fiducial points' are decided upon where each fiducial point represents a unique facial feature that will help in generating a representative face graph for that person.

For our analysis the fiducial point chosen were - Iris of left eye, Iris of right eye, nose tip, upper lip tip and chin tip.

Hence a labeled graph will have N = 5 nodes and E = 10 edges connecting between those points. Here an edge e<E connects two nodes n and n'.

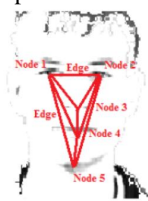

**Fig. 3.** The face graph that can be generated with the considered set of fiducial points

**Face Bunch Graph**

For large databases generating a separate face graph for each feature will create an excess database, which can be reduces by bunching data into a facial graph. Hence, a 'face bunch graph' is generated from the individual set that will have a stack of jets at each node that represents a fiducial point. [1]

Face bunch graph is generated for each individual person that represents uniquely the facial characteristics of that person. It has a set of jets from extracted from each model image representative of a particular person.

In our analysis, a small database is taken such that 3-4 face graphs of a person can successfully create a model bunch graph.

### 5.3 Training

After a bunch graph is generated representative of each individual, each input face segment is used to generate a face graph which is then iteratively matched with each and every of the bunch graph. [18]

Initially a training set of facial images is taken; each image is marked to a corresponding person. This set of image is used to manually generate face bunch graph [8]

For face recognition, fiducial point interior to the face region is important for identification procedure. For this, we choose out 5 fiducial points in the interior region of the face segment.

**Graph Similarity Matching**

Thus for an image graph $G$ with nodes n = 1, 2…N and edges e=1,2,…E matching is done between the corresponding parameters of the face bunch graph B as[18]:

$$S_B(G^I, B) = \frac{1}{N}\sum_n max_m \left(S_\phi(J_n^I, J_n^{Bm})\right) - \frac{\lambda}{E}\sum_e \frac{(\Delta \vec{x}_e^I - \Delta \vec{x}_e^B)^2}{(\Delta \vec{x}_e^B)^2} \quad (6)$$

Graph Similarity Measure between image graph and bunch graph

$\lambda$ decides the relative importance between jets and metric structure at 1 for our analysis. The locality about a fiducial point is taken to be of a range (+/- 5, +/-5) pixels.

### 5.4　Image Recognition

As in case of our analysis, there is face bunch graph representative of 5 test people. Each of the face bunch graphs is trained with a set of 2-3 image graphs [9].

Our recognizer matches the input image with the entire available face bunch graph and generates a unique similarity measure (Sb1, Sb2…Sb5).

Now this similarity measure value is input to the recognition thresholding limiter that generates binary output '1' or '0'. These outputs (O1, O2…O5) are multiplied by weights (W1, W2…W5) and then summed to generate a recognition index,

R=O1*W1+O2*W2+O3*W3+O4*W4+O5*W5 , **(7)**

Recognition Index for example database

If the Recognition index is:

1. '0', then the picture is not a match to the database
2. Same as any single weight(W1, W2,…W5), then it is perfect recognition
3. Sum of two or more weights, then it is over recognition requiring manual correction

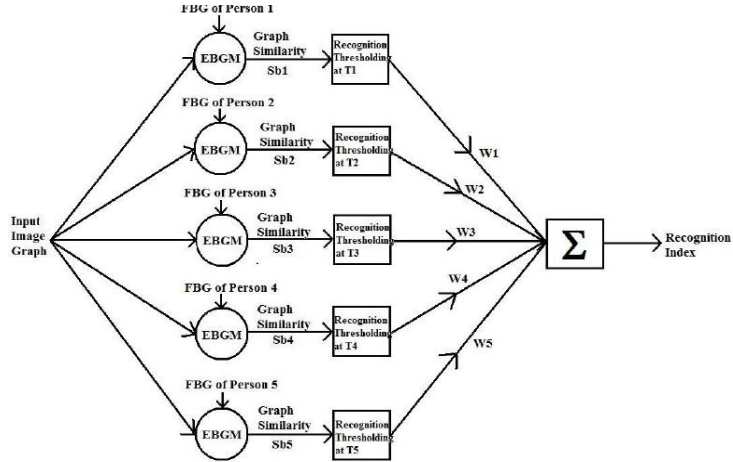

**Fig. 4.** Recognition network

## 6   Experimental Result

For result analysis, Face Recognition Database of University of Essex, UK is used which has color images of 395 individuals (male and female) with 20 images per individual of people of various races of age between 18 to 20 years.

To prove the utilization of the algorithm in real life applications where the availability of 20 images per person is improbable, each of the 15 individual Bunch Graph Models were trained using 1 image per individual with manually marked fiducial points.

For initialization of the algorithm, Similarity measure of each bunch graph is tested with 1 same individual image and 1 different individual image. For all the test images (arbitrarily selected) the similarity criteria for the similar images were found to be >0.9985 and that of the dissimilar images were found to be <0.9983 providing a threshold bandwidth of 0.0002.

The testing was performed on a truncated database of arbitrary 15 images per individual for all the 15 individuals. Therefore the total truncated database consisted of 225 facial images.

For best recognition rates the frequency of the Gabor Filters were restricted to $\gamma = \frac{\pi}{4}$ and the orientations were varied from $0, \frac{\pi}{8}, \ldots, \pi$.

**Table 1.** : Comparison of Matching Accuracy

| Matching Accuracy | Wiskott-EBGM | Skin-Segmentation followed by Wiskott EBGM |
|---|---|---|
| University of Essex Database | 85% | 88% |

## 7   Limitations

The Wiskott-EBGM is prone to translational and rotational distortions, and though the rotational distortions were considered zero degrees, the translation distortion was minimized to 5 pixels to reduce error margin, which reduces the robustness of the algorithm.

The Matching Rate was reduced due to over identification, which can be controlled for large databases using continuous user feedback.

## 8   Conclusion and future work

In this paper, a two-step facial recognition algorithm is implemented for analysis. Instead of directly implementing EBGM on images, EBGM algorithm is tested on skin segmented images. As the skin segmented images only have the necessary facial data, without background noises, it reduces the over identified Gabor features of the background and increases the efficiency.

The enhancement of the Matching Accuracy is only by 3% because though the background noise could be removed, the intra-facial region noises could not be eliminated. Also, as per our objective the modified two-step EBGM algorithm can successfully identify multiple facial regions in input images with multiple faces and extract them separately for matching.

For further enhancement of Matching Accuracy, the traditional Wiskott EBGM algorithm can be optimized to extract relevant facial features.

# 9    References

1. Beymer, D. (1994). Face recognition under varying pose. *Proc. IEEE Computer Vision and Pattern,* pp. 756-761.
2. Daugman, J. G. (1988). Complete discrete 2-D Gabor transform by neural networks for image analysis. *IEEE Trans. on Acoustics, Speech and Signal Processing.*
3. DeValois, R. L. (1988). *Spatial Vision.* Oxford Press.
4. Fleet, D. J. (1990). Computation of component image velocity from local phase information. *Int'l J. of Computer Vision,* 77-104.
5. Goldstein, A., Harmon, L., & Lesk, A. (May 1971). Identification of Human Faces. *IEEE Proceedings,* Vol. 59, No. 5, 748-760.
6. Gonzalez, R. C., & Woods, R. E. *Digital Image Processing.* New Jersey: Prentice Hall.
7. Jean-Marc Fellous, N. K. (1999). Face Recognition by Elastic Bunch Graph Matching. In L. Jain, *Intelligent Biometric Techniques in Fingerprint and Face Recognition* (pp. 355- 396). CRC Press.
8. Kruger, N. P. (1997). Determination of face position and. *Image and Vision Computing.*
9. Lades, M. V. (1993). Distortion invariant object recognition in the dynamic link architecture. *IEEE Trans. on Computers,* pp. 300-311.
10. Mohsin, W., Ahmed, N., & Mar, C.-T. (2003). *Face Detection Project.*
11. Movellan, J. *Introduction to Gabor Filters.* NSTC Subcommittee on Biometrics. (n.d.). Retrieved fromhttp://www.biometrics.gov/Documents/facerec.pdf
12. Pollen, D. A. (1981). *Phase relationship between adjacent simple cells in the visual cortex.* 1409-1411: Science.
13. Potzsch, M. K. (1997). Improving object recognition by transforming Gabor Filter responses. *Network: Computation in Neural Systems,* pp. 341-347.
14. Saveliev, P. (n.d.). *Euler Number - Computer Vision Primer.* Retrieved from Intelligent Perception: http://inperc.com/wiki/index.php?title=Euler_number
15. Sirovich, L., & Kirby, M. (1987). A Low-Dimensional procedure for Characterization of Human Faces. *Optical Soc. Am. A,* Vol. 4, No. 3, 519-524.
16. Theimer, W. M. (1994). Phase-based binocular vergence control and depth reconstruction using active vision. *CVGIP: Image Understanding,* 343-358.
17. Turk, M., & Pentland, A. (1991). Face Recognition Using Eigenfaces. *IEEE Proceedings,* 586-591.
18. Wiskott, L. F.-M. (1997). Face recognition by elastic bunch graph matching. *IEEE Trans. on Pattern Analysis and Machine Intelligence,* pp. 775-779.
19. Xiong, Z. *An Introduction to Face Detection and Recognition.*